\documentclass[runningheads]{llncs}
\usepackage{graphicx}
\usepackage{amsmath,amssymb} % define this before the line numbering.
\usepackage{color}
\usepackage[width=122mm,left=12mm,paperwidth=146mm,height=193mm,top=12mm,paperheight=217mm]{geometry}
\usepackage{hhline}

\usepackage{enumitem}
\usepackage{graphicx}
\usepackage{subfloat}
\usepackage{tabularx}
\usepackage{adjustbox}
\usepackage{amssymb}
\usepackage{multirow}
\usepackage[flushleft]{threeparttable}
\usepackage{subfigure}
\usepackage{tabularx}
\usepackage{rotating}
\usepackage{subfig}
\usepackage{blindtext}

\usepackage[T1]{fontenc}
\usepackage[utf8]{inputenc}
\usepackage{authblk}

\newcommand{\etal}{\textit{et al}. }
\newcommand{\ie}{\textit{i}.\textit{e}.}

\newcommand{\figref}[1]{Fig.~\ref{#1}}
\newcommand{\tabref}[1]{Table~\ref{#1}}

\newcommand{\sref}[1]{Sec.~\ref{#1}}
\newcommand*{\modulenamefull}{Convolutional Block Attention Module}
\newcommand*{\modulenameabb}{CBAM}
\begin{document}

\newcommand\blfootnote[1]{%
  \begingroup
  \renewcommand\thefootnote{}\footnote{#1}%
  \addtocounter{footnote}{-1}%
  \endgroup
}

\title{CBAM: \modulenamefull} 
% Replace with your title

\titlerunning{\modulenamefull}
% Replace with a meaningful short version of your title
%
\author{Sanghyun Woo\textsuperscript{*}\inst{1} 
\and
Jongchan Park{*$\dagger$}\inst{2} 
\and
Joon-Young Lee\inst{3}
\and
In So Kweon\inst{1}
}
%
%Please write out author names in full in the paper, i.e. full given and family names. 
%If any authors have names that can be parsed into FirstName LastName in multiple ways, please include the correct parsing, in a comment to the volume editors:
%\index{Lastnames, Firstnames}
%(Do not uncomment it, because you may introduce extra index items if you do that, we will use scripts for introducing index entries...)
\authorrunning{{Woo, Park, Lee, Kweon}}
% Replace with shorter version of the author list. If there are more authors than fits a line, please use A. Author et al.
%

\institute{Korea Advanced Institute of Science and Technology, Daejeon, Korea \\
\email{\{shwoo93, iskweon77\}@kaist.ac.kr}
\and
 Lunit Inc., Seoul, Korea \\
\email{jcpark@lunit.io}
\and
 Adobe Research, San Jose, CA, USA \\
 \email {jolee@adobe.com}
}

\maketitle

\blfootnote{*Both authors have equally contributed.}
\blfootnote{$\dagger$The work was done while the author was at KAIST.}

\vspace{-6mm}
\begin{abstract}
We propose \modulenamefull\ (\modulenameabb), a simple yet effective attention module for feed-forward convolutional neural networks. Given an intermediate feature map, our module sequentially infers attention maps along two separate dimensions, channel and spatial, then the attention maps are multiplied to the input feature map for adaptive feature refinement. %We place our attention module at every block of CNN models. 
Because \modulenameabb\ is a lightweight and general module, it can be integrated into any CNN architectures seamlessly with negligible overheads and is end-to-end trainable along with base CNNs. We validate our \modulenameabb\ through extensive experiments on ImageNet-1K, MS~COCO detection, and VOC~2007 detection datasets. Our experiments show consistent improvements in classification and detection performances with various models, demonstrating the wide applicability of \modulenameabb. The code and models will be publicly available.
% We propose Bottleneck Attention Module (BAM), a simple and effective attention module that can be integrated with any feed-forward convolutional neural networks. Given an intermediate feature map, our module infers an attention map along two separate pathways, channel and spatial, then the attention map is multiplied to the input feature map for adaptive feature refinement. We place our module at each `bottleneck' of models where the downsampling of feature maps occurs. Our module constructs a hierarchical attention at bottlenecks with a few number of parameters and it is trainable in an end-to-end manner jointly with any feed-forward models. We validate our BAM through extensive experiments on CIFAR-100, ImageNet-1K, and MS COCO detection datasets. Our experiments show consistent improvement on classification and detection performances with various models, demonstrating the wide applicability of BAM. The code and models will be publicly available upon the acceptance of this paper.
% \dots
\end{abstract}

\section{Introduction}

Convolutional neural networks (CNNs) have significantly pushed the performance of vision tasks~\cite{deng2009imagenet,krizhevsky2009learning,lin2014coco} based on their rich representation power. To enhance performance of CNNs, recent researches have mainly investigated three important factors of networks: \textit{depth}, \textit{width}, and \textit{cardinality}. 

From the LeNet architecture~\cite{lecun1998gradient} to Residual-style Networks~\cite{he2016deep,zagoruyko2016wide,xie2016aggregated,szegedy2017v4} so far, the network has become deeper for rich representation. VGGNet~\cite{simonyan2014very} shows that stacking blocks with the same shape gives fair results. Following the same spirit, ResNet~\cite{he2016deep} stacks the same topology of residual blocks along with skip connection to build an extremely deep architecture. %Based on the ResNet-style deep architecture, recent studies move on to investigate two different factors, width and cardinality. 
GoogLeNet \cite{szegedy2015going} shows that width is another important factor to improve the performance of a model. Zagoruyko and Komodakis~\cite{zagoruyko2016wide} propose to increase the width of a network based on the ResNet architecture. They have shown that a 28-layer ResNet with increased width can outperform an extremely deep ResNet with 1001 layers on the CIFAR benchmarks. Xception~\cite{chollet2016xception} and ResNeXt~\cite{xie2016aggregated} come up with to increase the cardinality of a network. They empirically show that cardinality not only saves the total number of parameters but also results in stronger representation power than the other two factors: depth and width.

\begin{figure}[t]
  \centering
  \includegraphics[width=\linewidth]{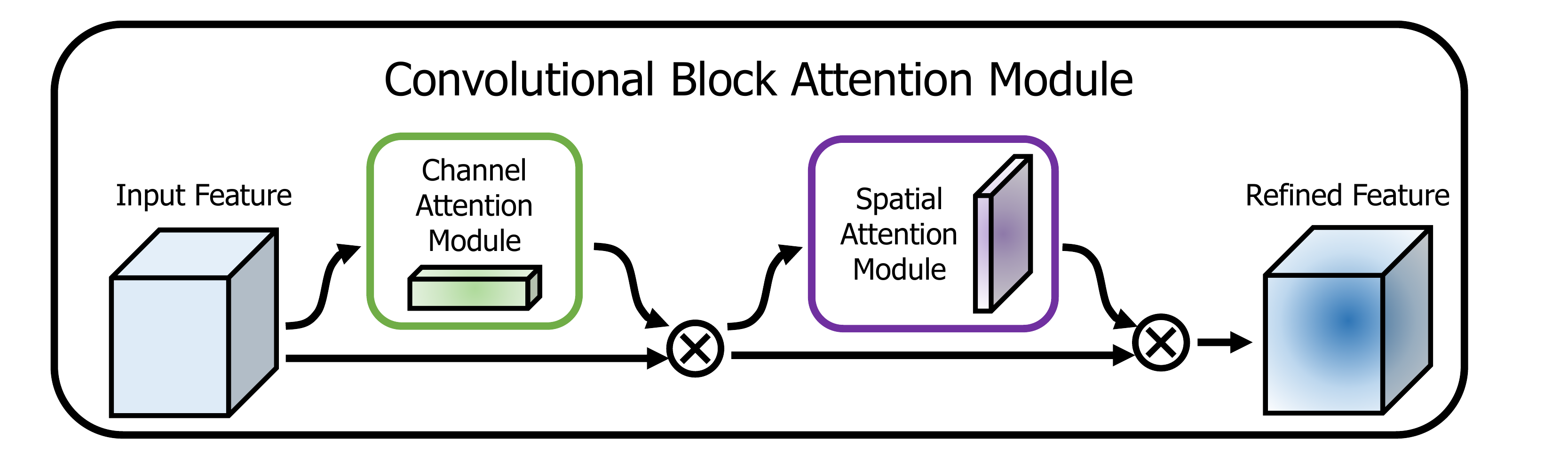}
  \caption{\textbf{The overview of CBAM}. The module has two sequential sub-modules: \textit{channel} and \textit{spatial}. The intermediate feature map is adaptively refined through our module (CBAM) at every convolutional block of deep networks.}
  \label{fig:overall}
  \vspace{-6mm}
\end{figure}

Apart from these factors, we investigate a different aspect of the architecture design, \textit{attention}. The significance of attention has been studied extensively in the previous literature~\cite{mnih2014recurrent,ba2014multiple,Bahdanau2014,xu2015show,gregor2015draw,Jaderberg2015}. 
%Attention not only provides where to focus but also intensifies diverse representations of objects at that location. 
Attention not only tells where to focus, it also improves the representation of interests.
Our goal is to increase representation power by using attention mechanism: focusing on important features and suppressing unnecessary ones. In this paper, we propose a new network module, named ``\modulenamefull''. Since convolution operations extract informative features by blending cross-channel and spatial information together, we adopt our module to emphasize meaningful features along those two principal dimensions: channel and spatial axes. To achieve this, we sequentially apply channel and spatial attention modules (as shown in \figref{fig:overall}), so that each of the branches can learn `what' and `where' to attend in the channel and spatial axes respectively. As a result, our module efficiently helps the information flow within the network by learning which information to emphasize or suppress.

In the ImageNet-1K dataset, we obtain accuracy improvement from various baseline networks by plugging our tiny module, revealing the efficacy of \modulenameabb. 
We visualize trained models using the grad-CAM~\cite{selvaraju2017grad} and observe that \modulenameabb-enhanced networks focus on target objects more properly than their baseline networks.
% Interestingly, we have observed that multiple \modulenameabb\ located at different convolution blocks build a hierarchical attention as shown in \figref{fig:teaser}. 
Taking this into account, we conjecture that the performance boost comes from accurate attention and noise reduction of irrelevant clutters. Finally, we validate performance improvement of object detection on the MS COCO and the VOC 2007 datasets, demonstrating a wide applicability of \modulenameabb. Since we have carefully designed our module to be light-weight, the overhead of parameters and computation is negligible in most cases.

\smallskip\noindent\textbf{Contribution.} Our main contribution is three-fold.
\begin{enumerate}[topsep=0pt,itemsep=0pt]
\item We propose a simple yet effective attention module (\modulenameabb) that can be widely applied to boost representation power of CNNs.
%We propose a simple and effective attention module (BAM) which shows a wide applicability for boosting representation power of CNNs.
\item We validate the effectiveness of our attention module through extensive ablation studies.
\item We verify that performance of various networks is greatly improved on the multiple benchmarks (ImageNet-1K, MS COCO, and VOC 2007) by plugging our light-weight module.
%We verify that BAM outperforms all the baselines of various networks on the benchmarks (CIFAR-100, ImageNet-1K, and MS COCO) without bells and whistles.
\end{enumerate}

\section{Related Work}\label{sec:related}

\subsubsection{Network engineering.} 
``Network engineering'' has been one of the most important vision research, because well-designed networks ensure remarkable performance improvement in various applications. A wide range of architectures has been proposed since the successful implementation of a large-scale CNN~\cite{krizhevsky2012imagenet}. An intuitive and simple way of extension is to increase the depth of neural networks~\cite{simonyan2014very}. Szegedy~\etal~\cite{szegedy2015going} introduce a deep Inception network using a multi-branch architecture where each branch is customized carefully. While a naive increase in depth comes to saturation due to the difficulty of gradient propagation, ResNet~\cite{he2016deep} proposes a simple identity skip-connection to ease the optimization issues of deep networks. Based on the ResNet architecture, various models such as WideResNet~\cite{zagoruyko2016wide}, Inception-ResNet~\cite{szegedy2017v4}, and ResNeXt~\cite{xie2016aggregated} have been developed. WideResNet~\cite{zagoruyko2016wide} proposes a residual network with a larger number of convolutional filters and reduced depth. PyramidNet~\cite{han2016deep} is a strict generalization of WideResNet where the width of the network gradually increases. ResNeXt~\cite{xie2016aggregated} suggests to use grouped convolutions and shows that increasing the cardinality leads to better classification accuracy. More recently, Huang~\etal~\cite{huang2016densely} propose a new architecture, DenseNet. It iteratively concatenates the input features with the output features, enabling each convolution block to receive raw information from all the previous blocks. While most of recent network engineering methods mainly target on three factors \textit{depth}~\cite{krizhevsky2012imagenet,simonyan2014very,szegedy2015going,he2016deep}, \textit{width}~\cite{szegedy2015going,szegedy2016rethinking,zagoruyko2016wide,szegedy2017v4}, and \textit{cardinality}~\cite{xie2016aggregated,chollet2016xception}, we focus on the other aspect, `\textit{attention}', one of the curious facets of a human visual system.

\subsubsection{Attention mechanism.}
It is well known that attention plays an important role in human perception~\cite{Itti1998saliency,rensink2000dynamic,corbetta2002control}. One important property of a human visual system is that one does not attempt to process a whole scene at once. Instead, humans exploit a sequence of partial glimpses and selectively focus on salient parts in order to capture visual structure better~\cite{larochelle2010learning}. 

Recently, there have been several attempts~\cite{wang2017residual,hu2017squeeze} to incorporate attention processing to improve the performance of CNNs in large-scale classification tasks. Wang~\etal\cite{wang2017residual} propose \textit{Residual Attention Network} which uses an encoder-decoder style attention module. By refining the feature maps, the network not only performs well but is also robust to noisy inputs. Instead of directly computing the 3d attention map, we decompose the process that learns channel attention and spatial attention separately. The separate attention generation process for 3D feature map has much less computational and parameter overhead, and therefore can be used as a plug-and-play module for pre-existing base CNN architectures.
% We adopt the dilated convolution~\cite{yu2015multi} for the simplicity rather than the encoder-decoder style architecture. 

More close to our work, Hu~\etal~\cite{hu2017squeeze} introduce a compact module to exploit the inter-channel relationship. In their \textit{Squeeze-and-Excitation} module, they use global average-pooled features to compute channel-wise attention. However, we show that those are suboptimal features in order to infer fine channel attention, and we suggest to use max-pooled features as well. They also miss the spatial attention, which plays an important role in deciding `where' to focus as shown in \cite{chen2016sca}. In our \modulenameabb, we exploit both spatial and channel-wise attention based on an efficient architecture and empirically verify that exploiting both is superior to using only the channel-wise attention as \cite{hu2017squeeze}.
Moreover, we empirically show that our module is effective in detection tasks~(MS-COCO and VOC). Especially, we achieve state-of-the-art performance just by placing our module on top of the existing one-shot detector~\cite{woo2017stairnet} in the VOC2007 test set.

\section {\modulenamefull}

\begin{figure*}[t]
  \centering
  \includegraphics[width=\linewidth]{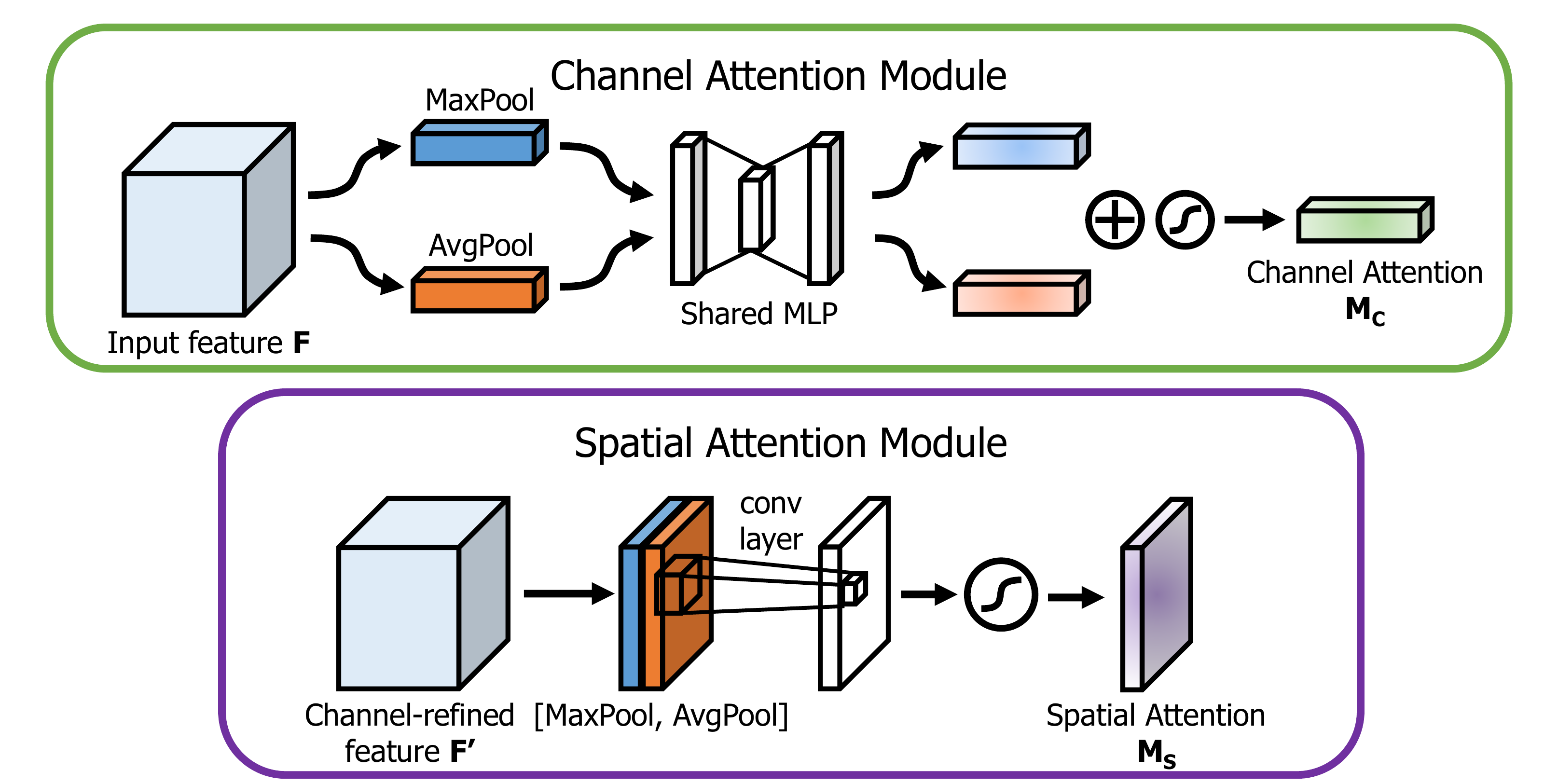}
  \caption{\textbf{Diagram of each attention sub-module.} As illustrated, the channel sub-module utilizes both max-pooling outputs and average-pooling outputs with a shared network; the spatial sub-module utilizes similar two outputs that are pooled along the channel axis and forward them to a convolution layer.}
  \label{fig:teaser}
  \vspace{-6mm}
\end{figure*}

%%% Method %%%
Given an intermediate feature map \(\mathbf{F}\in \mathbb{R}^{C\times H\times W}\) as input, \modulenameabb\ sequentially infers a 1D channel attention map \(\mathbf{M_c}\in \mathbb{R}^{C\times 1\times 1}\) and a 2D spatial attention map \(\mathbf{M_s}\in \mathbb{R}^{1\times H\times W}\) as illustrated in \figref{fig:overall}.
% The refined feature map $\mathbf{F}'$ is computed as:
% \begin{equation}\label{eq:first}
%     \mathbf{F}'=\mathbf{F}+ \mathbf{F} \otimes \mathbf{M}(\mathbf{F}),
% \end{equation}
% where $\otimes$ denotes element-wise multiplication.
% We adopt a residual attention mechanism to facilitate the gradient flow.
%We adopt an attention mechanism along with the residual learning scheme to facilitate the gradient flow. 
% To design an efficient yet powerful module, we first compute the channel attention \(\mathbf{M_c}(\mathbf{F})\in \mathbb{R}^{C}\) and then the spatial attention \(\mathbf{M_s}(\mathbf{F})\in \mathbb{R}^{H\times W}\) sequentially. 
The overall attention process can be summarized  as:
%the attention map \(\mathbf{M}(\mathbf{F})\) is computed as a combination of two separate branches; the channel branch \(\mathbf{M_c}(\mathbf{F})\in \mathbb{R}^{C}\), and the spatial branch \(\mathbf{M_s}(\mathbf{F})\in \mathbb{R}^{H\times W}\):
\begin{equation}\label{eq:first}
    % \mathbf{M}(\mathbf{F})=\mathbf{M_s}(\mathbf{M_c}(\mathbf{F})),
\begin{split}
    \mathbf{F'}&=\mathbf{M_c}(\mathbf{F})  \otimes \mathbf{F}, \\
    \mathbf{F''}&=\mathbf{M_s}(\mathbf{F'}) \otimes \mathbf{F'},
\end{split}
\end{equation}
where $\otimes$ denotes element-wise multiplication. During multiplication, the attention values are broadcasted (copied) accordingly: channel attention values are broadcasted along the spatial dimension, and vice versa. \(\mathbf{F''}\) is the final refined output. \figref{fig:teaser} depicts the computation process of each attention map. The following describes the details of each attention module.

\subsubsection{Channel attention module.}
We produce a channel attention map by exploiting the inter-channel relationship of features. 
% The channel attention values, ranging from 0 to 1, are multiplied to the input feature map to emphasize or suppress the channels.
As each channel of a feature map is considered as a feature detector~\cite{zeiler2014visualizing}, channel attention focuses on `what' is meaningful given an input image. To compute the channel attention efficiently, we squeeze the spatial dimension of the input feature map.
% The channel attention map is \textit{global}, as a single channel attention value is applied over all the spatial locations in the channel. 
% To aggregate global statistics over the spatial dimension, we squeeze the spatial dimension of the input feature map. 
% Using the aggregated global feature, the computation of channel attention is much efficient. 
For aggregating spatial information, average-pooling has been commonly adopted so far. Zhou~\etal~\cite{zhou2016learning} suggest to use it to learn the extent of the target object effectively and Hu~\etal~\cite{hu2017squeeze} adopt it in their attention module to compute spatial statistics.
% Intuitively, it is a simple way to aggregate the spatial statistics and has a dense gradient update in an one-shot manner.
Beyond the previous works, we argue that max-pooling gathers another important clue about distinctive object features to infer finer channel-wise attention. Thus, we use both average-pooled and max-pooled features simultaneously. We empirically confirmed that exploiting both features greatly improves representation power of networks rather than using each independently (see \sref{sec:ablation}), showing the effectiveness of our design choice. 
We describe the detailed operation below. 

We first aggregate spatial information of a feature map by using both average-pooling and max-pooling operations, generating two different spatial context descriptors: \(\mathbf{F^{c}_{avg}}\) and \(\mathbf{F^{c}_{max}}\), which denote average-pooled features and max-pooled features respectively. Both descriptors are then forwarded to a shared network to produce our channel attention map \(\mathbf{M_c}\in \mathbb{R}^{C\times 1\times 1}\). The shared network is composed of multi-layer perceptron (MLP) with one hidden layer. To reduce parameter overhead, the hidden activation size is set to \(\mathbb{R}^{C/r\times 1\times 1}\), where \(r\) is the reduction ratio. After the shared network is applied to each descriptor, we merge the output feature vectors using element-wise summation. In short, the channel attention is computed as:
\begin{equation}\label{eq:third}
\begin{split}
    \mathbf{M_c}(\mathbf{F})&=\sigma(MLP(AvgPool(\mathbf{F}))+MLP(MaxPool(\mathbf{F})))\\
    &=\sigma( \mathbf{W_1}(\mathbf{W_0}(\mathbf{F^{c}_{avg}}))+
    \mathbf{W_1}(\mathbf{W_0}(\mathbf{F^{c}_{max}}))),
\end{split}
\end{equation}
where \(\sigma\) denotes the sigmoid function, \(\mathbf{W_0}\in \mathbb{R}^{C/r\times C}\), and \(\mathbf{W_1}\in \mathbb{R}^{C\times C/r}\). Note that the MLP weights, \(\mathbf{W_0}\) and \(\mathbf{W_1}\), are shared for both inputs and the ReLU activation function is followed by \(\mathbf{W_0}\).

\subsubsection{Spatial attention module.}
We generate a spatial attention map by utilizing the inter-spatial relationship of features. Different from the channel attention, the spatial attention focuses on `where' is an informative part, which is complementary to the channel attention. To compute the spatial attention, we first apply average-pooling and max-pooling operations along the channel axis and concatenate them to generate an efficient feature descriptor. Applying pooling operations along the channel axis is shown to be effective in highlighting informative regions~\cite{Zagoruyko2017AT}. %Also, it makes the spatial submodule symmetrical with our channel attention computation. 
On the concatenated feature descriptor, we apply a convolution layer to generate a spatial attention map \(\mathbf{M_s}(\mathbf{F})\in \mathbf{R}^{H\times W}\) which encodes where to emphasize or suppress. We describe the detailed operation below.

We aggregate channel information of a feature map by using two pooling operations, generating two 2D maps: \(\mathbf{F^{s}_{avg}}\in \mathbb{R}^{1\times H\times W}\) and \(\mathbf{F^{s}_{max}}\in \mathbb{R}^{1\times H\times W}\). Each denotes average-pooled features and max-pooled features across the channel. Those are then concatenated and convolved by a standard convolution layer, producing our 2D spatial attention map. In short, the spatial attention is computed as:
\begin{equation}\label{eq:forth}
\begin{split}
    \mathbf{M_s}(\mathbf{F})&=\sigma(f^{7\times 7}([AvgPool(\mathbf{F}); MaxPool(\mathbf{F})]))\\
    &=\sigma(f^{7\times 7}([\mathbf{F^{s}_{avg}}; \mathbf{F^{s}_{max}}])),
\end{split}
\end{equation}
where \(\sigma\) denotes the sigmoid function and \(f^{7\times 7}\) represents a convolution operation with the filter size of $7\times 7$.

\subsubsection{Arrangement of attention modules.}
Given an input image, two attention modules, channel and spatial, compute complementary attention, focusing on `what' and `where' respectively. Considering this, two modules can be placed in a parallel or sequential manner. We found that the sequential arrangement gives a better result than a parallel arrangement. For the arrangement of the sequential process, our experimental result shows that the channel-first order is slightly better than the spatial-first. We will discuss experimental results on network engineering in \sref{sec:ablation}. %This indicates that sequentially-generated 3D attention map in either way is an effective approximation of the ideal attention map.

\section {Experiments}\label{sec:experiments}
We evaluate CBAM on the standard benchmarks: ImageNet-1K for image classification; MS COCO and VOC 2007 for object detection. In order to perform better apple-to-apple comparisons, we reproduced all the evaluated networks~\cite{he2016deep,zagoruyko2016wide,xie2016aggregated,howard2017mobilenets,hu2017squeeze} in the PyTorch framework~\cite{pytorch} and report our reproduced results in the whole experiments.

\begin{figure*}[t]
  \centering
  \includegraphics[width=\linewidth]{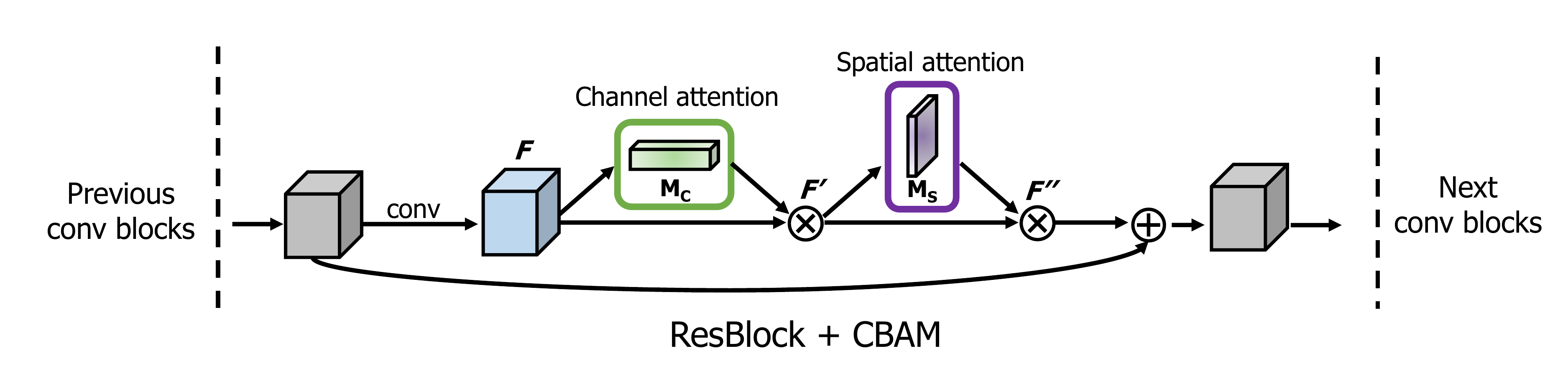}
  \caption{\textbf{CBAM integrated with a ResBlock in ResNet\cite{he2016deep}.} This figure shows the exact position of our module when integrated within a ResBlock. We apply CBAM on the convolution outputs in each block.}
  \label{fig:convblock}
  \vspace{-6mm}
\end{figure*}

To thoroughly evaluate the effectiveness of our final module, we first perform extensive ablation experiments. Then, we verify that \modulenameabb\ outperforms all the baselines without bells and whistles, demonstrating the general applicability of \modulenameabb\ across different architectures as well as different tasks. One can seamlessly integrate CBAM in any CNN architectures and jointly train the combined CBAM-enhanced networks. \figref{fig:convblock} shows a diagram of CBAM integrated with a ResBlock in ResNet~\cite{he2016deep} as an example.

\subsection{Ablation studies}
\label{sec:ablation}
In this subsection, we empirically show the effectiveness of our design choice. For this ablation study, we use the ImageNet-1K dataset and adopt ResNet-50~\cite{he2016deep} as the base architecture. The ImageNet-1K classification dataset~\cite{deng2009imagenet} consists of 1.2 million images for training and 50,000 for validation with 1,000 object classes. We adopt the same data augmentation scheme with \cite{he2016deep,he2016identity} for training and apply a single-crop evaluation with the size of 224$\times$224 at test time. The learning rate starts from 0.1 and drops every 30 epochs. We train the networks for 90 epochs. Following \cite{he2016deep,he2016identity,huang2016deep}, we report classification errors on the validation set.

Our module design process is split into three parts. We first search for the effective approach to computing the channel attention, then the spatial attention. Finally, we consider how to combine both channel and spatial attention modules. We explain the details of each experiment below.

\subsubsection{Channel attention.}\label{sec:channel_ablation}

%% Tabel 1, Ablation Study 1 : channel ablation 
\begin{table}[b]
\begin{center}
\begin{adjustbox}{max width=\textwidth}
\begin{tabular}{ l|c|c|c|c }
\hline
Description & Parameters & GFLOPs & Top-1 Error(\%) & Top-5 Error(\%) \\
\hline
\hline
ResNet50 (baseline)                           & 25.56M                & 3.86          
                                                    & 24.56   &   7.50\\
ResNet50 + AvgPool (SE~\cite{hu2017squeeze})    & 25.92M    & 3.94
                                                    & 23.14  &   6.70\\
\hline
ResNet50 + MaxPool                            & 25.92M    & 3.94  
                                                    & 23.20  &   6.83\\
ResNet50 + AvgPool \& MaxPool                 & 25.92M    & 4.02  
                                                    &  \textbf{22.80}  &   \textbf{6.52}\\
\hline
\end{tabular}
\end{adjustbox}
\end{center}
\caption{\textbf{Comparison of different channel attention methods}. We observe that using our proposed method outperforms recently suggested Squeeze and Excitation method~\cite{hu2017squeeze}.} 
\label{table:Ablation_channel}
\end{table}

We experimentally verify that using both average-pooled and max-pooled features enables finer attention inference. We compare 3 variants of channel attention: average pooling, max pooling, and joint use of both poolings. Note that the channel attention module with an average pooling is the same as the SE~\cite{hu2017squeeze} module. Also, when using both poolings, we use a shared MLP for attention inference to save parameters, as both of aggregated channel features lie in the same semantic embedding space. We only use channel attention modules in this experiment and we fix the reduction ratio to 16.

Experimental results with various pooling methods are shown in \tabref{table:Ablation_channel}. We observe that max-pooled features are as meaningful as average-pooled features, comparing the accuracy improvement from the baseline. In the work of SE~\cite{hu2017squeeze}, however, they only exploit the average-pooled features, missing the importance of max-pooled features. We argue that max-pooled features which encode the degree of the most salient part can compensate the average-pooled features which encode global statistics softly. Thus, we suggest to use both features simultaneously and apply a shared network to those features. The outputs of a shared network are then merged by element-wise summation.  We empirically show that our channel attention method is an effective way to push performance further from SE~\cite{hu2017squeeze} without additional learnable parameters. As a brief conclusion, we use both average- and max-pooled features in our channel attention module with the reduction ratio of 16 in the following experiments.

%% Tabel 2, Ablation Study 2 : spatial ablation 
\begin{table}[t]
\begin{center}
\begin{adjustbox}{max width=\textwidth}
\begin{tabular}{ l|c|c|c|c }
\hline
Description &  Param. &  GFLOPs &  Top-1 Error(\%) &  Top-5 Error(\%) \\
\hline
\hline
ResNet50 + channel (SE~\cite{hu2017squeeze})                            & 28.09M    & 3.860
                                                    & 23.14 &   6.70\\
\hline
ResNet50 + channel                            & 28.09M    & 3.860
                                                    & 22.80  &   6.52\\
ResNet50 + channel + spatial (1x1 conv, k=3)  & 28.10M & 3.862
                                                    & 22.96  &   6.64\\
ResNet50 + channel + spatial (1x1 conv, k=7)  & 28.10M & 3.869
                                                    & 22.90  &   6.47\\
ResNet50 + channel + spatial (avg\&max, k=3)  & 28.09M & 3.863
                                                    & 22.68  &   6.41\\
ResNet50 + channel + spatial (avg\&max, k=7)  & 28.09M & 3.864
                                                    & \textbf{22.66}  &   \textbf{6.31}\\
\hline
\end{tabular}
\end{adjustbox}
\end{center}
\caption{\textbf{Comparison of different spatial attention methods}. Using the proposed channel-pooling (\ie\ average- and max-pooling along the channel axis) along with the large kernel size of 7 for the following convolution operation performs best.}
\label{table:Ablation_spatial}
\end{table}

%% Tabel 2, Ablation Study 2 : spatial ablation 
\begin{table}[t]
\begin{center}
\begin{adjustbox}{max width=\textwidth}
\begin{tabular}{l|c|c }
\hline
Description & Top-1 Error(\%) & Top-5 Error(\%) \\
\hline
\hline
ResNet50 + channel (SE~\cite{hu2017squeeze})
            & 23.14 &   6.70\\
\hline
ResNet50 + channel + spatial                  & \textbf{22.66}    & \textbf{6.31} \\
ResNet50 + spatial + channel                  & 22.78    & 6.42\\
ResNet50 + channel \& spatial in parallel     & 22.95        & 6.59\\
\hline
\end{tabular}
\end{adjustbox}
\end{center}
\caption{\textbf{Combining methods of channel and spatial attention}. Using both attention is critical while the best-combining strategy (\ie\ sequential, channel-first) further improves the accuracy.}
\vspace{-6mm}
\label{table:Ablation_sequence}
\end{table}

\subsubsection{Spatial attention.}
Given the channel-wise refined features, we explore an effective method to compute the spatial attention. The design philosophy is symmetric with the channel attention branch. To generate a 2D spatial attention map, we first compute a 2D descriptor that encodes channel information at each pixel over all spatial locations. We then apply one convolution layer to the 2D descriptor, obtaining the raw attention map. The final attention map is normalized by the sigmoid function. 

We compare two methods of generating the 2D descriptor: \textit{channel pooling} using average- and max-pooling across the channel axis and \textit{standard \(1\times 1\) convolution} reducing the channel dimension into 1. In addition, we investigate the effect of a kernel size at the following convolution layer: kernel sizes of 3 and 7. In the experiment, we place the spatial attention module after the previously designed channel attention module, as the final goal is to use both modules together.

\tabref{table:Ablation_spatial} shows the experimental results. We can observe that the channel pooling produces better accuracy, indicating that explicitly modeled pooling leads to finer attention inference rather than learnable weighted channel pooling (implemented as \(1\times 1\) convolution). In the comparison of different convolution kernel sizes, we find that adopting a larger kernel size generates better accuracy in both cases. It implies that a broad view (\ie\ large receptive field) is needed for deciding spatially important regions. Considering this, we adopt the channel-pooling method and the convolution layer with a large kernel size to compute spatial attention. In a brief conclusion, we use the average- and max-pooled features across the channel axis with a convolution kernel size of 7 as our spatial attention module.

\subsubsection{Arrangement of the channel and spatial attention.}
In this experiment, we compare three different ways of arranging the channel and spatial attention submodules: sequential channel-spatial, sequential spatial-channel, and parallel use of both attention modules. As each module has different functions, the order may affect the overall performance. For example, from a spatial viewpoint, the channel attention is globally applied, while the spatial attention works locally. Also, it is natural to think that we may combine two attention outputs to build a 3D attention map. In the case, both attentions can be applied in parallel, then the outputs of the two attention modules are added and normalized with the sigmoid function.

\tabref{table:Ablation_sequence} summarizes the experimental results on different attention arranging methods. From the results, we can find that generating an attention map sequentially infers a finer attention map than doing in parallel. In addition, the channel-first order performs slightly better than the spatial-first order. Note that all the arranging methods outperform using only the channel attention independently, showing that utilizing both attentions is crucial while the best-arranging strategy further pushes performance.

\subsubsection{Final module design.}
Throughout the ablation studies, we have designed the channel attention module, the spatial attention module, and the arrangement of the two modules. Our final module is as shown in \figref{fig:overall} and \figref{fig:teaser}: we choose average- and max-pooling for both channel and spatial attention module; we use convolution with a kernel size of 7 in the spatial attention module; we arrange the channel and spatial submodules sequentially. Our final module(\ie\ ResNet50 + CBAM) achieves top-1 error of 22.66\%, which is much lower than SE~\cite{hu2017squeeze}(\ie\ ResNet50 + SE), as shown in \tabref{table:imagenet_exp_1}.

%% Tabel 2, ImageNet1000 Evaluation Results
\begin{table}[t]
\begin{center}
%\resizebox{0.8\textwidth}{!}{%
\begin{adjustbox}{max width=\textwidth}
\begin{tabular}{l|c|c|c|c }
\hline
Architecture  & Param. & GFLOPs & Top-1 Error (\%) & Top-5 Error  (\%) \\
\hline
\hline
ResNet18 \cite{he2016deep}                                  & 11.69M & 1.814 & 29.60 &10.55 \\
ResNet18 \cite{he2016deep} + SE~\cite{hu2017squeeze}         & 11.78M & 1.814 & 29.41 & 10.22\\
ResNet18 \cite{he2016deep} + \modulenameabb                 & 11.78M & 1.815
                                                            & \textbf{29.27} &\textbf{10.09}\\
\hline
ResNet34 \cite{he2016deep}                                  & 21.80M & 3.664 & 26.69 & 8.60 \\
ResNet34 \cite{he2016deep} + SE~\cite{hu2017squeeze}         & 21.96M & 3.664 & 26.13 & 8.35  \\
ResNet34 \cite{he2016deep} + \modulenameabb                 & 21.96M & 3.665
                                                            &\textbf{25.99} & \textbf{8.24}  \\
\hline
ResNet50 \cite{he2016deep}                                  & 25.56M & 3.858 &24.56 & 7.50 \\
ResNet50 \cite{he2016deep} + SE~\cite{hu2017squeeze}         & 28.09M & 3.860 & 23.14 & 6.70  \\
ResNet50 \cite{he2016deep} + \modulenameabb                 & 28.09M & 3.864
                                                            &\textbf{22.66} & \textbf{6.31}  \\
\hline
ResNet101 \cite{he2016deep}                                 & 44.55M & 7.570 & 23.38 & 6.88 \\
ResNet101 \cite{he2016deep} + SE~\cite{hu2017squeeze}        & 49.33M & 7.575 & 22.35 & 6.19 \\
ResNet101 \cite{he2016deep} + \modulenameabb                & 49.33M & 7.581
                                                            &\textbf{21.51} & \textbf{5.69} \\

\hline
WideResNet18 \cite{zagoruyko2016wide} (widen=1.5)                           & 25.88M & 3.866 & 26.85 & 8.88 \\
WideResNet18 \cite{zagoruyko2016wide} (widen=1.5) + SE~\cite{hu2017squeeze} & 26.07M & 3.867 & 26.21 & 8.47 \\
WideResNet18 \cite{zagoruyko2016wide} (widen=1.5) + \modulenameabb          & 26.08M & 3.868 & \textbf{26.10}& \textbf{8.43} \\
\hline

WideResNet18 \cite{zagoruyko2016wide} (widen=2.0)                           & 45.62M & 6.696 & 25.63 & 8.20 \\
WideResNet18 \cite{zagoruyko2016wide} (widen=2.0) + SE~\cite{hu2017squeeze} & 45.97M & 6.696 & 24.93 & 7.65 \\
WideResNet18 \cite{zagoruyko2016wide} (widen=2.0) + \modulenameabb          & 45.97M & 6.697 & \textbf{24.84}& \textbf{7.63} \\

% WideResNet50 \cite{zagoruyko2016wide} (widen=1.5)               & - & - & - & \\
% WideResNet50 \cite{zagoruyko2016wide} (widen=1.5) + SE~\cite{hu2017squeeze} & - & - & - & - \\
% WideResNet50 \cite{zagoruyko2016wide} (widen=1.5) + \modulenameabb & - & - & - & - \\
\hline
ResNeXt50 \cite{xie2016aggregated} (32x4d)                              & 25.03M & 3.768 & 22.85  & 6.48 \\
ResNeXt50 \cite{xie2016aggregated} (32x4d) + SE~\cite{hu2017squeeze}    & 27.56M & 3.771 & \textbf{21.91}  & 6.04 \\
ResNeXt50 \cite{xie2016aggregated} (32x4d) + \modulenameabb             & 27.56M & 3.774 & 21.92 & \textbf{5.91}\\
\hline

ResNeXt101 \cite{xie2016aggregated} (32x4d)                             & 44.18M & 7.508 & 21.54 & 5.75 \\
ResNeXt101 \cite{xie2016aggregated} (32x4d) + SE~\cite{hu2017squeeze}   & 48.96M & 7.512 & 21.17 & 5.66 \\
ResNeXt101 \cite{xie2016aggregated} (32x4d) + \modulenameabb            & 48.96M & 7.519 & \textbf{21.07} & \textbf{5.59} \\
\hline
\end{tabular}
\end{adjustbox}
%}
\end{center}
\begin{tablenotes}
\scriptsize
\item \vspace{-3mm}\hspace*{\fill} \textbf{*} all results are reproduced in the PyTorch framework.
\end{tablenotes}
\caption{ \textbf{Classification results on ImageNet-1K.} Single-crop validation errors are reported.}
\label{table:imagenet_exp_1}
\vspace{-6mm}
\end{table}

\begin{figure}[t]
\begin{tabular}{cc}
\includegraphics[width=.5\linewidth]{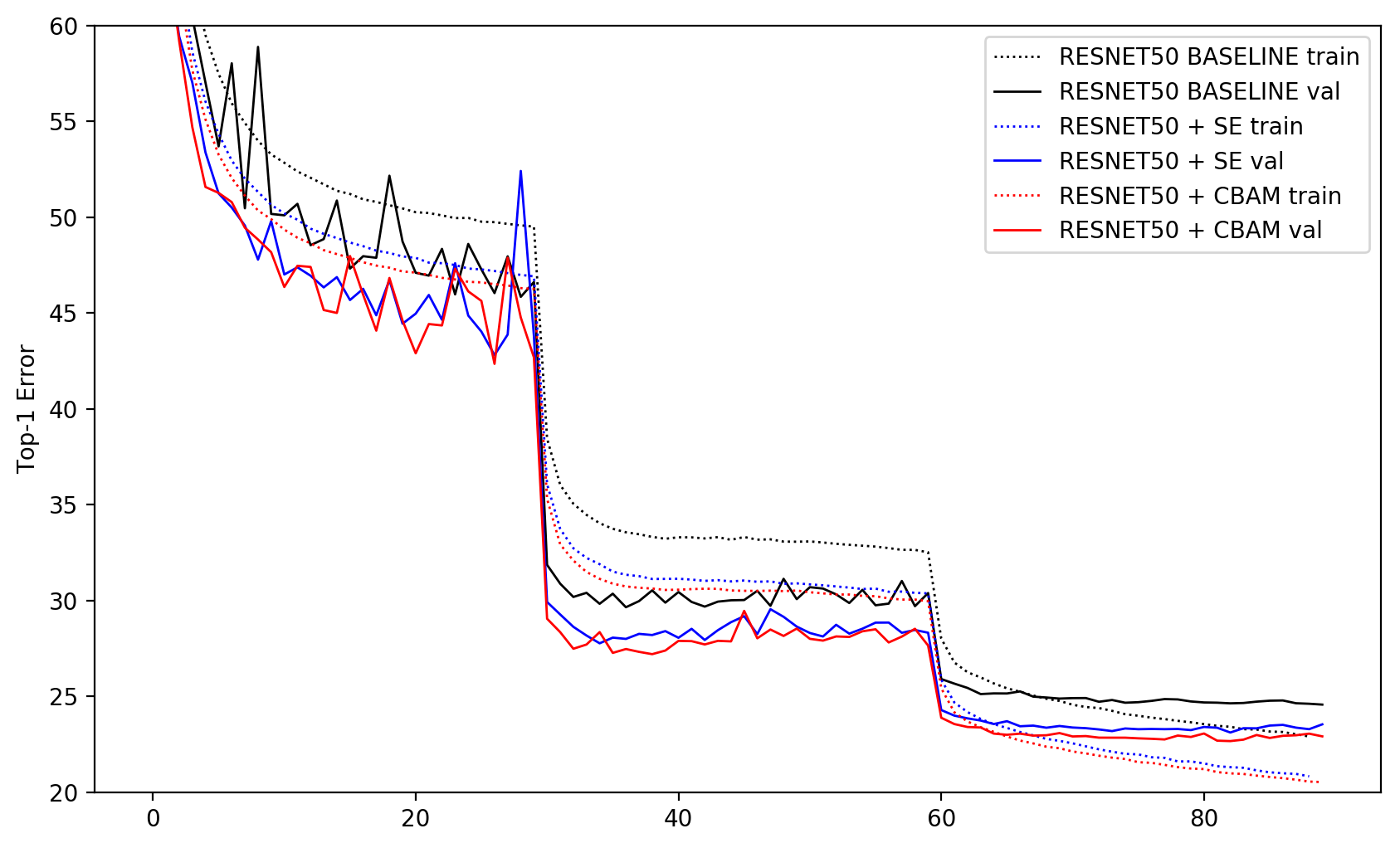}& \includegraphics[width=.5\linewidth]{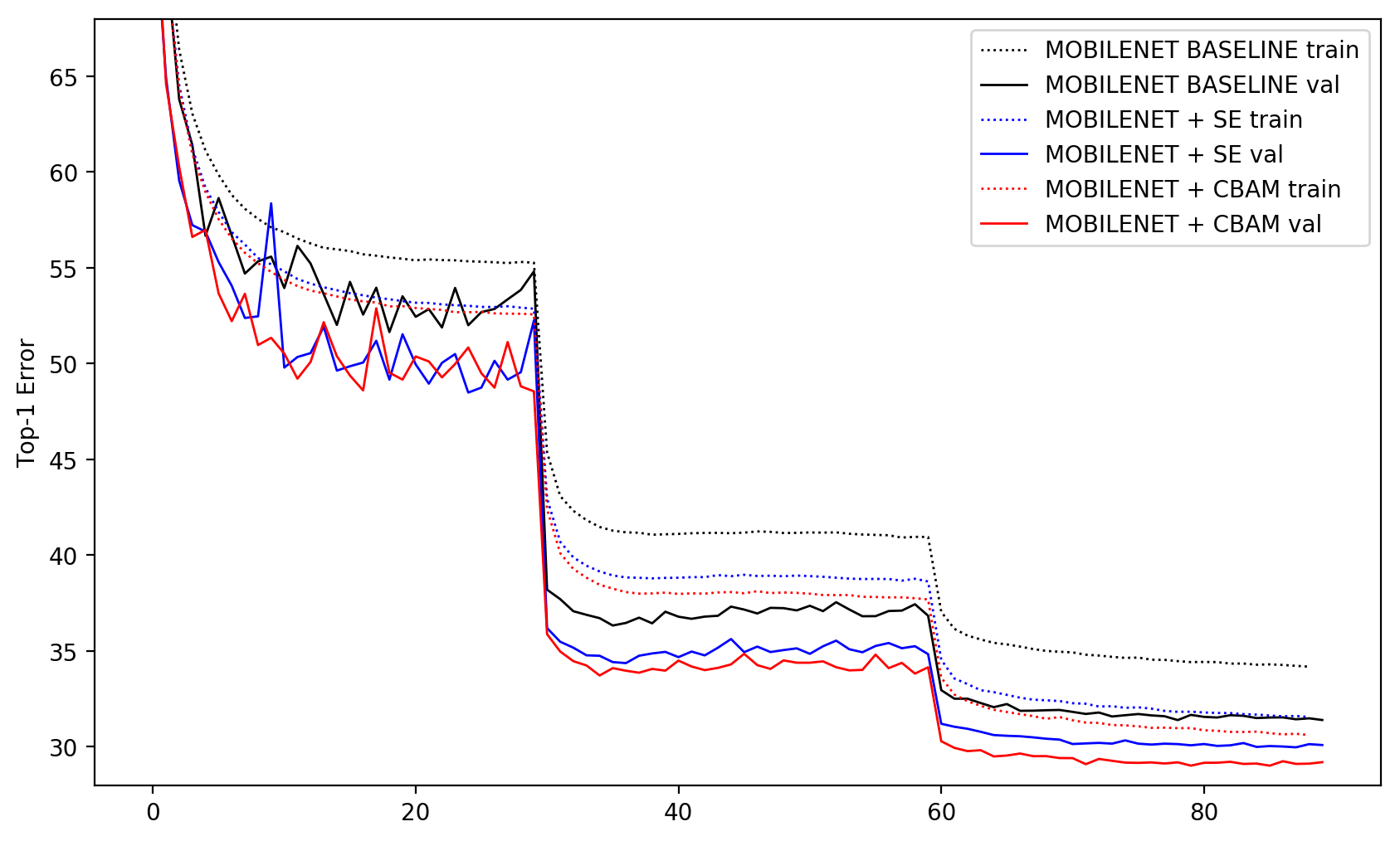} \\
\small (a) ResNet50~\cite{he2016deep} & \small (b) MobileNet~\cite{howard2017mobilenets}
\end{tabular}\vspace{2mm}
\caption{\textbf{Error curves during ImageNet-1K training}. Best viewed in color.}
\label{fig:resnet50_plot}
\end{figure}

\begin{table}[t]
\begin{center}
%\resizebox{0.8\textwidth}{!}{%
\begin{adjustbox}{max width=\textwidth}
\begin{tabular}{l|c|c|c|c }
\hline
Architecture  & Parameters & GFLOPs & Top-1 Error (\%) & Top-5 Error (\%) \\
\hline\hline
MobileNet \cite{howard2017mobilenets} \(\alpha=0.7\)                           & 2.30M & 0.283 & 34.86 & 13.69 \\
MobileNet \cite{howard2017mobilenets} \(\alpha=0.7\) + SE~\cite{hu2017squeeze} & 2.71M & 0.283 & 32.50 & 12.49 \\
MobileNet \cite{howard2017mobilenets} \(\alpha=0.7\) + \modulenameabb          & 2.71M & 0.289 & \textbf{31.51} & \textbf{11.48} \\
\hline
MobileNet \cite{howard2017mobilenets}                           & 4.23M & 0.569 & 31.39 & 11.51 \\
MobileNet \cite{howard2017mobilenets} + SE~\cite{hu2017squeeze} & 5.07M & 0.570 & 29.97 & 10.63 \\
MobileNet \cite{howard2017mobilenets} + \modulenameabb          & 5.07M & 0.576 & \textbf{29.01} & \textbf{9.99} \\
\hline
\end{tabular}
\end{adjustbox}
\end{center}
\begin{tablenotes}
\scriptsize
\item \vspace{-3mm}\hspace*{\fill} \textbf{*} all results are reproduced in the PyTorch framework.
\end{tablenotes}\vspace{2mm}
\caption{ \textbf{Classification results on ImageNet-1K using the light-weight network, MobileNet~\cite{howard2017mobilenets}.} Single-crop validation errors are reported.}
\label{table:imagenet_exp_2}
\vspace{-6mm}
\end{table}

\subsection{Image Classification on ImageNet-1K}

% As shown in \tabref{table:imagenet_exp}, the networks with BAM outperform all the baselines once again, demonstrating that the BAM can generalize well on various models in the large-scale dataset. Note that the overhead of parameters and computation is negligible, which suggests that the proposed module BAM can significantly enhance the network capacity efficiently. Another notable thing is that the improved performance comes from placing only three modules overall the network.

We perform ImageNet-1K classification experiments to rigorously evaluate our module. We follow the same protocol as specified in \sref{sec:ablation} 
and evaluate our module in various network architectures including ResNet~\cite{he2016deep}, WideResNet~\cite{zagoruyko2016wide}, and ResNext~\cite{xie2016aggregated}.

\tabref{table:imagenet_exp_1} summarizes the experimental results. The networks with CBAM outperform all the baselines significantly, demonstrating that the CBAM can generalize well on various models in the large-scale dataset. 
Moreover, the models with CBAM improve the accuracy upon the one of the strongest method -- SE~\cite{hu2017squeeze} which is the winning approach of the ILSVRC 2017 classification task. 
It implies that our proposed approach is powerful, showing the efficacy of \textit{new pooling method} that generates richer descriptor and \textit{spatial attention} that complements the channel attention effectively. 

\figref{fig:resnet50_plot} depicts the error curves of various networks during ImageNet-1K training. We can clearly see that our method exhibits lowest training and validation error in both error plots. It shows that CBAM has greater ability to improve generalization power of baseline models compared to SE~\cite{hu2017squeeze}.

We also find that the overall overhead of CBAM is quite small in terms of both parameters and computation. This motivates us to apply our proposed module \modulenameabb\ 
to the light-weight network, MobileNet~\cite{howard2017mobilenets}. 
\tabref{table:imagenet_exp_2} summarizes the experimental results that we conducted based on the MobileNet architecture. 
We have placed CBAM to two models, basic and capacity-reduced model(\ie\ adjusting width multiplier(\(\alpha\)) to 0.7). 
We observe similar phenomenon as shown in \tabref{table:imagenet_exp_1}. CBAM not only boosts the accuracy of baselines significantly but also favorably improves the performance of SE~\cite{hu2017squeeze}. This shows the great potential of CBAM for applications on low-end devices.

\begin{figure*}
  \centering
  \includegraphics[width=\linewidth]{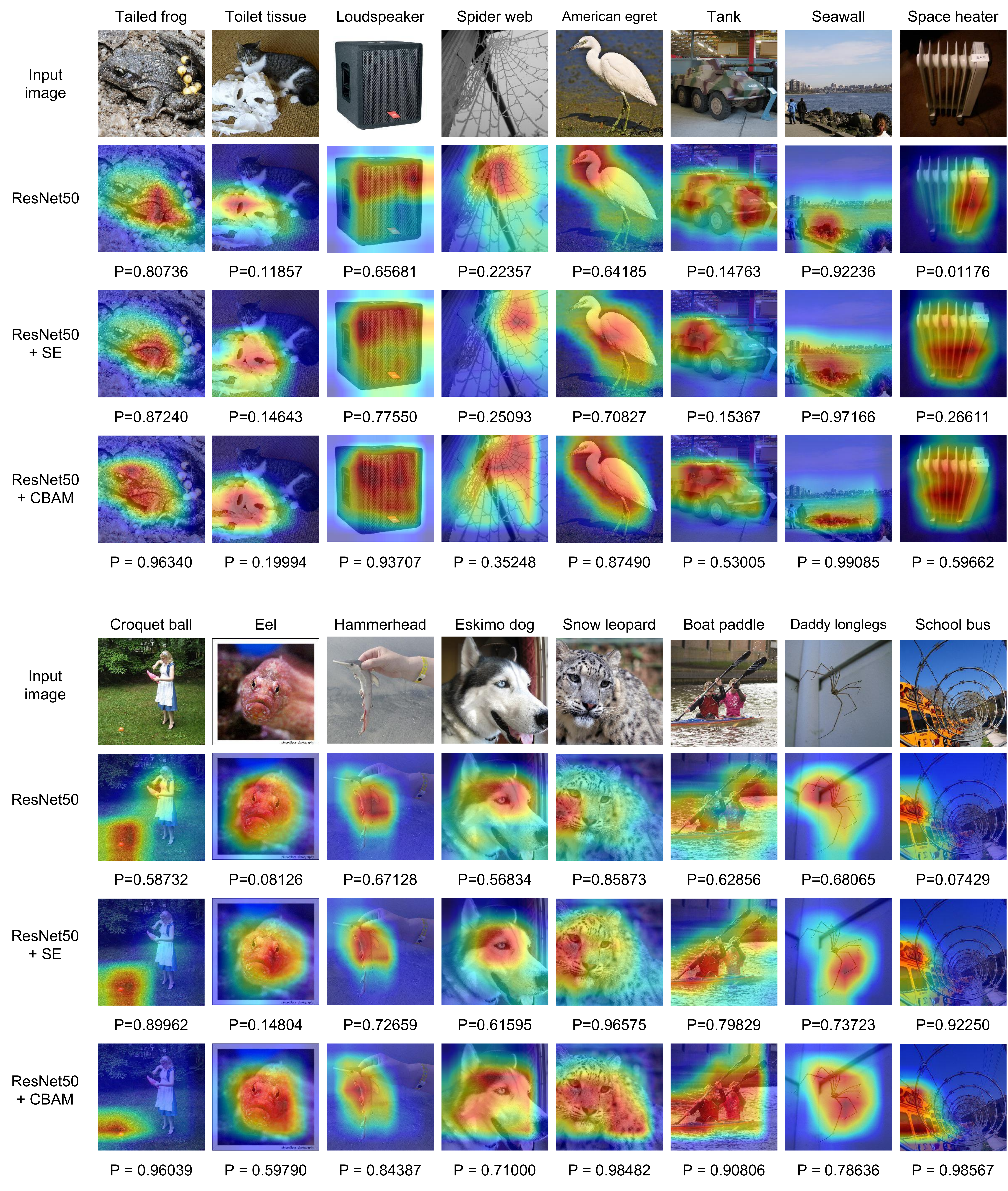}
  \caption{\textbf{Grad-CAM~\cite{selvaraju2017grad} visualization results.} We compare the visualization results of CBAM-integrated network (ResNet50 + CBAM) with baseline (ResNet50) and SE-integrated network (ResNet50 + SE). The grad-CAM visualization is calculated for the last convolutional outputs. The ground-truth label is shown on the top of each input image and \textit{P} denotes the softmax score of each network for the ground-truth class.}
  \label{fig:grad_cam}
\end{figure*}

\subsection{Network Visualization with Grad-CAM~\cite{selvaraju2017grad}}

For the qualitative analysis, we apply the Grad-CAM~\cite{selvaraju2017grad} to different networks using images from the ImageNet validation set. Grad-CAM is a recently proposed visualization method which uses gradients in order to calculate the importance of the spatial locations in convolutional layers. As the gradients are calculated with respect to a unique class, Grad-CAM result shows attended regions clearly. By observing the regions that network has considered as important for predicting a class, we attempt to look at how this network is making good use of features. We compare the visualization results of CBAM-integrated network (ResNet50 + CBAM) with baseline (ResNet50) and SE-integrated network (ResNet50 + SE). \figref{fig:grad_cam} illustrate the visualization results. The softmax scores for a target class are also shown in the figure. 

In \figref{fig:grad_cam}, we can clearly see that the Grad-CAM masks of the CBAM-integrated network cover the target object regions better than other methods. That is, the CBAM-integrated network learns well to exploit information in target object regions and aggregate features from them. Note that target class scores also increase accordingly. From the observations, we conjecture that the feature refinement process of CBAM eventually leads the networks to utilize given features well.

% aggregating features from the target object's spatial location. \figref{fig:grad_cam} shows several samples of Grad-CAM visualization results. We compare the baseline ResNet101, ResNet101 with SE\cite{hu2017squeeze}, and ResNet101 with CBAM. 
% We can observe that ResNet101 with CBAM tends to capture features exclusively from target object locations. That is, CBAM helps the network to aggregate features spatially.

\subsection{MS COCO Object Detection}

We conduct object detection on the Microsoft COCO dataset~\cite{lin2014coco}. This dataset involves 80k training images (``2014 train'') and 40k validation images (``2014 val''). The average mAP over different IoU thresholds from 0.5 to 0.95 is used for evaluation. According to \cite{bell2016ion,liu2016ssd}, we trained our model using all the training images as well as a subset of validation images, holding out 5,000 examples for validation. Our training code is based on \cite{chen17implementation} and we train the network for 490K iterations for fast performance validation. We adopt \textit{Faster-RCNN}~\cite{ren2015faster} as our detection method and ImageNet pre-trained ResNet50 and ResNet101~\cite{he2016deep} as our baseline networks. Here we are interested in performance improvement by plugging CBAM to the baseline networks. Since we use the same detection method in all the models, the gains can only be attributed to the enhanced representation power, given by our module CBAM. As shown in the \tabref{table:coco_detect}, we observe significant improvements from the baseline, demonstrating generalization performance of CBAM on other recognition tasks.

\begin{table}[t]
\begin{center}
\setlength{\tabcolsep}{2pt}
\begin{tabular}{ l|l|c|c|c}
\hline
Backbone &Detector            & mAP@.5 & mAP@.75 & mAP@[.5, .95] \\
\hline
\hline
ResNet50~\cite{he2016deep} & Faster-RCNN~\cite{ren2015faster}      & 46.2           & 28.1          & 27.0     \\
ResNet50~\cite{he2016deep} + \modulenameabb  & Faster-RCNN~\cite{ren2015faster}      & \textbf{48.2}  & \textbf{29.2} & \textbf{28.1}     \\
\hline
ResNet101~\cite{he2016deep} &Faster-RCNN~\cite{ren2015faster}      & 48.4           & 30.7          & 29.1     \\
ResNet101~\cite{he2016deep} + \modulenameabb  &Faster-RCNN~\cite{ren2015faster}     & \textbf{50.5}  & \textbf{32.6} & \textbf{30.8}     \\
\hline
\end{tabular}
\end{center}
\begin{tablenotes}
\scriptsize
\item \vspace{-3mm}\hspace*{\fill} \textbf{*} all results are reproduced in the PyTorch framework.
\end{tablenotes}\vspace{2mm}
\caption{\textbf{Object detection mAP(\%) on the MS COCO validation set}. We adopt the Faster R-CNN~\cite{ren2015faster} detection framework and apply our module to the base networks. CBAM boosts mAP@[.5,~.95] by 0.9 for both baseline networks.}
\label{table:coco_detect}
\vspace{-6mm}
\end{table}

\begin{table}[t]
\begin{center}
\setlength{\tabcolsep}{2pt}
\begin{tabular}{ l|l|c|c}
\hline
Backbone & Detector & mAP@.5 & Parameters (M) \\
\hline
\hline
VGG16~\cite{simonyan2014very} & SSD~\cite{liu2016ssd}        &  77.8  & 26.5\\
VGG16~\cite{simonyan2014very} &
StairNet~\cite{woo2017stairnet} & 78.9 & 32.0\\
VGG16~\cite{simonyan2014very} &
StairNet~\cite{woo2017stairnet} + SE~\cite{hu2017squeeze} & 79.1 & 32.1\\
VGG16~\cite{simonyan2014very} &
StairNet~\cite{woo2017stairnet} + CBAM & \textbf{79.3} & 32.1\\
\hline
MobileNet~\cite{howard2017mobilenets} & SSD~\cite{liu2016ssd}        &  68.1   & 5.81\\
MobileNet~\cite{howard2017mobilenets} &
StairNet~\cite{woo2017stairnet} & 70.1 & 5.98\\
MobileNet~\cite{howard2017mobilenets} &
StairNet~\cite{woo2017stairnet} + SE~\cite{hu2017squeeze} & 70.0 & 5.99\\
MobileNet~\cite{howard2017mobilenets} &
StairNet~\cite{woo2017stairnet} + CBAM                   & \textbf{70.5} &6.00\\
\hline
\end{tabular}
\end{center}
\begin{tablenotes}
\scriptsize
\item \vspace{-3mm}\hspace*{\fill} \textbf{*} all results are reproduced in the PyTorch framework.
\end{tablenotes}\vspace{2mm}
\caption{\textbf{Object detection mAP(\%) on the VOC 2007 test set}. We adopt the StairNet~\cite{woo2017stairnet} detection framework and apply SE and CBAM to the detectors. CBAM favorably improves all the strong baselines with negligible additional parameters.}
\label{table:voc_detect}
\vspace{-6mm}
\end{table}

\subsection{VOC 2007 Object Detection}

We further perform experiments on the PASCAL VOC 2007 test set. In this experiment, we apply CBAM to the detectors, while the previous experiments (\tabref{table:coco_detect}) apply our module to the base networks. We adopt the StairNet~\cite{woo2017stairnet} framework, which is one of the strongest multi-scale method based on the SSD~\cite{liu2016ssd}. For the experiment, we reproduce SSD and StairNet in our PyTorch platform in order to estimate performance improvement of CBAM accurately and achieve 77.8\% and 78.9\% mAP@.5 respectively, which are higher than the original accuracy reported in the original papers. We then place SE~\cite{hu2017squeeze} and CBAM right before every classifier, refining the final features which are composed of up-sampled global features and corresponding local features before the prediction, enforcing model to adaptively select only the meaningful features. We train all the models on the union set of VOC 2007 trainval and VOC 2012 trainval (``07+12''), and evaluate on the VOC 2007 test set. The total number of training epochs is 250. We use a weight decay of 0.0005 and a momentum of 0.9.
In all the experiments, the size of the input image is fixed to 300 for the simplicity. 

The experimental results are summarized in \tabref{table:voc_detect}. We can clearly see that CBAM improves the accuracy of all strong baselines with two backbone networks. Note that accuracy improvement of CBAM comes with a negligible parameter overhead, indicating that enhancement is not due to a naive capacity-increment but because of our effective feature refinement. In addition, the result using the light-weight backbone network~\cite{howard2017mobilenets} again shows that CBAM can be an interesting method to low-end devices.

\section{Conclusion}

We have presented the convolutional bottleneck attention module (CBAM), a new approach to improve representation power of CNN networks. 
We apply attention-based feature refinement with two distinctive modules, channel and spatial, and achieve considerable performance improvement while keeping the overhead small.
For the channel attention, we suggest to use the max-pooled features along with the average-pooled features, leading to produce finer attention than SE~\cite{hu2017squeeze}. We further push the performance by exploiting the spatial attention.
Our final module (CBAM) learns what and where to emphasize or suppress and refines intermediate features effectively.
% Inspired by a human visual system, we suggest placing an attention module at the bottleneck of a network where the most critical points of information flow.
%We enjoy the performance improvement only by plugging our light-weight, two-pathway module which learns what and where to focus or suppress. 
%Inspired by a human visual system which exploits information bottleneck mechanism, we place BAM at each bottleneck of a network. 
To verify its efficacy, we conducted extensive experiments with various state-of-the-art models and confirmed that CBAM outperforms all the baselines on three different benchmark datasets: ImageNet-1K, MS COCO, and VOC 2007. In addition, we visualize how the module exactly infers given an input image. Interestingly, we observed that our module induces the network to focus on target object properly. We hope CBAM become an important component of various network architectures.

\clearpage

\bibliographystyle{splncs}
\bibliography{egbib}
\end{document}